\pdfoutput=1

\documentclass[journal,transmag]{IEEEtran}
\ifCLASSINFOpdf
   \usepackage[pdftex]{graphicx}
\else
\usepackage[dvips]{graphicx}
\fi
\usepackage{times}
\usepackage{latexsym}
\usepackage{enumitem}
\usepackage{tabulary}
\usepackage{adjustbox}
\usepackage{booktabs}
\usepackage{float}
\usepackage{caption}
\usepackage{multirow}
\usepackage{multicol}
\usepackage{tcolorbox}
\usepackage[utf8]{inputenc}

\usepackage{hyperref}

\newcommand{\comb}[1]{\textcolor{blue}{#1}}

\usepackage{pifont}
\newcommand{\cmark}{\ding{51}}%
\newcommand{\xmark}{\ding{55}}%

\usepackage[T1]{fontenc}

\usepackage[utf8]{inputenc}

\usepackage{microtype}

\usepackage{cite}
\usepackage{amsmath}
\ifCLASSOPTIONcompsoc
 \usepackage[caption=false,font=normalsize,labelfont=sf,textfont=sf]{subfig}
\else
 \usepackage[caption=false,font=footnotesize]{subfig}
\fi
\usepackage{url}

\begin{document}
%
\title{Unsupervised Domain Adaptation with Global and Local Graph Neural Networks in Limited Labeled Data Scenario \\ {\large Application to Disaster Management}}


\author{\IEEEauthorblockN{Samujjwal Ghosh\IEEEauthorrefmark{1},
Subhadeep Maji\IEEEauthorrefmark{2}, and 
Maunendra Sankar Desarkar\IEEEauthorrefmark{1}, 
}
\IEEEauthorblockA{\IEEEauthorrefmark{1}Indian Institute of Technology Hyderabad, India}
\IEEEauthorblockA{\IEEEauthorrefmark{2}Amazon, India}
}

\markboth{GLEN}%
{Ghosh \MakeLowercase{\textit{et al.}}: Bare Demo of IEEEtran.cls for IEEE Transactions on Magnetics Journals}
%



\IEEEtitleabstractindextext{%
\begin{abstract}
Identification and categorization of social media posts generated during disasters are crucial to reduce the sufferings of the affected people. However, lack of labeled data is a significant bottleneck in learning an effective categorization system for a disaster. This motivates us to study the problem as unsupervised domain adaptation (UDA) between a previous disaster with labeled data (source) and a current disaster (target). However, if the amount of labeled data available is limited, it restricts the learning capabilities of the model. To handle this challenge, we utilize limited labeled data along with abundantly available unlabeled data, generated during a source disaster to propose a novel two-part graph neural network. The first-part extracts domain-agnostic global information by constructing a token level graph across domains and the second-part preserves local instance-level semantics. In our experiments, we show that the proposed method outperforms state-of-the-art techniques by $2.74\%$ weighted F$_1$ score on average on two standard public dataset in the area of disaster management. We also report experimental results for granular actionable multi-label classification datasets in disaster domain for the first time, on which we outperform BERT by $3.00\%$ on average w.r.t weighted F$_1$. Additionally, we show that our approach can retain performance when very limited labeled data is available. We plan to release our code publicly.
\end{abstract}
\begin{IEEEkeywords}
Domain Adaptation, Graph Neural Networks, Text Classification, Disaster Management
\end{IEEEkeywords}}

\maketitle

\IEEEdisplaynontitleabstractindextext

%
\IEEEpeerreviewmaketitle

\section{Introduction}
%
%
%
%
Identifying relevant and actionable information from the data generated during a disaster can help mitigate the sufferings of the affected people in those areas. During and after a disaster, lots of social media posts are generated from the affected regions expressing the condition and the difficulties faced by the people. On the other hand, several individuals, government and non-government agencies post about their arrangements, activities and intents towards extending help in the rescue and relief operations. Timely extraction and identification of such actionable information from social media can be beneficial for disaster relief~\cite{dis_www, taqe, 8955890}. 
Table~\ref{table:example_tweet} presents an example tweet posted during the 2015 Nepal earthquake and the different actionable classes that it corresponds to. The example indicates that the generated posts can contain useful, actionable information, and can be of multi-label in nature.
\begin{table}[b]
\centering
\begin{adjustbox}{totalheight=1.8\height,
}
\begin{tabular}{p{1cm}p{1.3cm}|p{1.3cm}|p{1.2cm}| p{1.2cm}}
 \toprule
 \multicolumn{5}{c}{\textbf{Tweet}} \\
 \midrule 
 \multicolumn{5}{c}{\shortstack{two C130 J and two C17 aircraft will be \\ in Nepal today carrying rescue teams, doctors, \\ field mobile hospitals, sniffer dogs, blankets, food}} \\
 \midrule
  & \shortstack{Resource \\ Available} & \shortstack{Resource \\ Required} & \shortstack{Infra \\ Damage} & \shortstack{Govt. \\ Activities} \\ \cmidrule(l){2-5}
 \textbf{Label} & \cmark & \xmark &\cmark & \xmark \\
 \bottomrule
\end{tabular}
\end{adjustbox}
\caption{\small{An example tweet from FIRE16 dataset involving 2015 Nepal earthquake. The labels represent actionable disaster response, and multiple labels are applicable for the tweet.}}
\label{table:example_tweet}
\end{table}
However, a large portion of generated data are sentiments of people towards the disaster. Filtering and categorizing the small percentage of actionable information from this large pool is challenging. One primary reason for this is the difficulty in obtaining a sufficiently large number of annotations on time to assist development of a classifier. The problem has been explored in literature \cite{dis_www, ghosh2020semi} with the assumption that sufficient amount of labeled data is available. However, this assumption might not be realistic during a disaster.

In this work, we focus on the problem of identification and categorization of actionable social media posts when no labeled data is available from the current disaster. Unsupervised Domain Adaptation (UDA) may be utilized in such a scenario if labeled data from a prior disaster is available \cite{sigir'20, qcri, caragea2016identifying}.
However, even when labeled data from a past disaster is available, the amount of labeled data is generally minimal as annotating posts based on the actionable information is expensive. Lack of labeled data might restrict learning a classifier for any effective representation of domain shift.

To overcome these challenges, we propose an UDA approach based on a cross-domain graph construction which utilizes abundantly available unlabeled data from both past (source) and current (target) disaster.
To this effect, we propose a novel two-part (\textit{Global} and \textit{Local}) graph-based formalism. The \textit{Global}-part learns a domain-agnostic feature representation for tokens using a large token graph constructed from labeled data from source domain along with unlabeled data from both the domains to identify analogous terms between the domains and brings them closer. The \textit{Local}-part operates per instance, preserves the local instance-level semantics by learning contextual representations of the posts. We combine these representations and learn a classifier that generalizes across domains. We refer to our approach as \textbf{G}lobal and \textbf{L}ocal Graph N\textbf{e}ural \textbf{N}etwork (\textbf{GLEN}).

In our experiments, we compare GLEN against SOTA methods for UDA in the area of disaster management. We report results on datasets consisting of tweets involving binary-classification where GLEN is better than SOTA methods by $3.74\%$ on average. In real-world multi-label and multi-class classification setting, we show that GLEN is better than BERT on average. To the best of our knowledge, our work is the first attempt to study UDA on multi-label datasets for disaster management.
We also show GLEN outperforms BERT by an even larger-margin when both are trained on a smaller fraction of labeled data, making GLEN more attractive in domains with small amount of labeled data.
Taking a step further, we also conduct some qualitative investigation to understand the capabilities and limitations of GLEN better. We show that GLEN is indeed able to learn domain agnostic representations for tokens. It does so by bringing tokens which appear in similar context across the two domains nearer to each other in the embedding space. In summary, we make the following contributions in the present work. 

\begin{itemize}[leftmargin=*]
\setlength{\itemsep}{0pt}
\item We effectively utilize the unlabeled data from both source and target domains for UDA in the area of disaster management with a novel two-part graph neural network based framework.
\item We investigate the multi-label setting in UDA for the first time in disaster management and show that our model performs competitively in the multi-label setting as well. 
\item We show significant improvement for UDA over SOTA methods and BERT on a number of datasets. We achieve on average $3.74\%$ improvement over SOTA methods and $1.21\%$ improvement over BERT.
\end{itemize}

\section{Related Work} \label{sec:related_work}
\subsection{Unsupervised Domain Adaptation (UDA)} \label{sec:related_uda} UDA is an ideal task for disaster management due to the inherent lack of labeled data in target domain. Existing UDA approaches can be categorized into four major types:
\begin{itemize}
    \item Feature projection \cite{chen2012marginalized, AdapEn, sigir'20}
    \item Instance re-weighting \cite{ijcai2018-0624, cui2019self, iscram}
    \item Pivot feature centric \cite{ziser-reichart-2018-pivot, ziser2019task}
    \item Domain Adversarial / Gradient Reversal based \cite{wu2019domain, qcri, adasd, adadqd, adamrc, dis_image}
\end{itemize}
\textbf{Feature projection} signifies bringing the features of source and target domain to a joint latent space. \cite{chen2012marginalized} used stacked autoencoders to learn domain adaptive feature representations for sentiment analysis.
In \textbf{instance re-weighting}, source domain samples are identified which are similar to target domain, and the model is re-trained with weighted instances from source domain. However, \cite{negative_re-weighting} showed that instance re-weighting might suffer from poor performance if there is limited similarity between the domains. In such a scenario, the model assigns large weights to a few samples resulting in a bias towards those samples.
This limitation is addressed in \cite{ijcai2018-0624} by introducing \textit{sample selection variance} in instance re-weighting. Sample selection variance ensures that weights are distributed over a larger population of source data and do not rely on a few highly relevant instances. Instance re-weighting and feature projection is combined in \cite{ijcai2018-0624}. Features are projected such that class differences are maximized, which reduces misclassification. Pseudo labels are then learned for unlabeled target samples. Finally, the classifier is updated using the target domain data along with their pseudo labels as true classes. Instance re-weighting approaches are computationally expensive as the model needs to be re-trained.
\textbf{Pivot Based} Language Modelling (PBLM) approaches work by predicting if a word is pivot between source and target domains. \cite{ziser-reichart-2018-pivot} applied PBLM on sentiment classification task. However, finding a large number of pivots is challenging. To mitigate the challenge, the authors proposed Task Refinement Learning (TRL-PBLM) in their followup work \cite{ziser2019task} by dividing the pivots into subsets and predicting the subset instead of the pivot directly.
\textbf{Domain Adversarial} approaches are frequent for (un)supervised domain adaptation. \cite{wu2019domain, adasd, adadqd, adamrc} use gradient reversal for news classification, stance detection, duplicate question detection and machine reading comprehension respectively. Main objective of these approaches is to maximize domain difference using a gradient reversal layer over domain classification loss. \cite{wu2019domain} uses Domain-Adversarial Graph Neural Network (DAGNN) to represent both source and target domain documents as instance graphs to capture local context only in both domains.
%
%
\begin{figure*}[t]
    \centering
    \includegraphics[width=1.\linewidth]{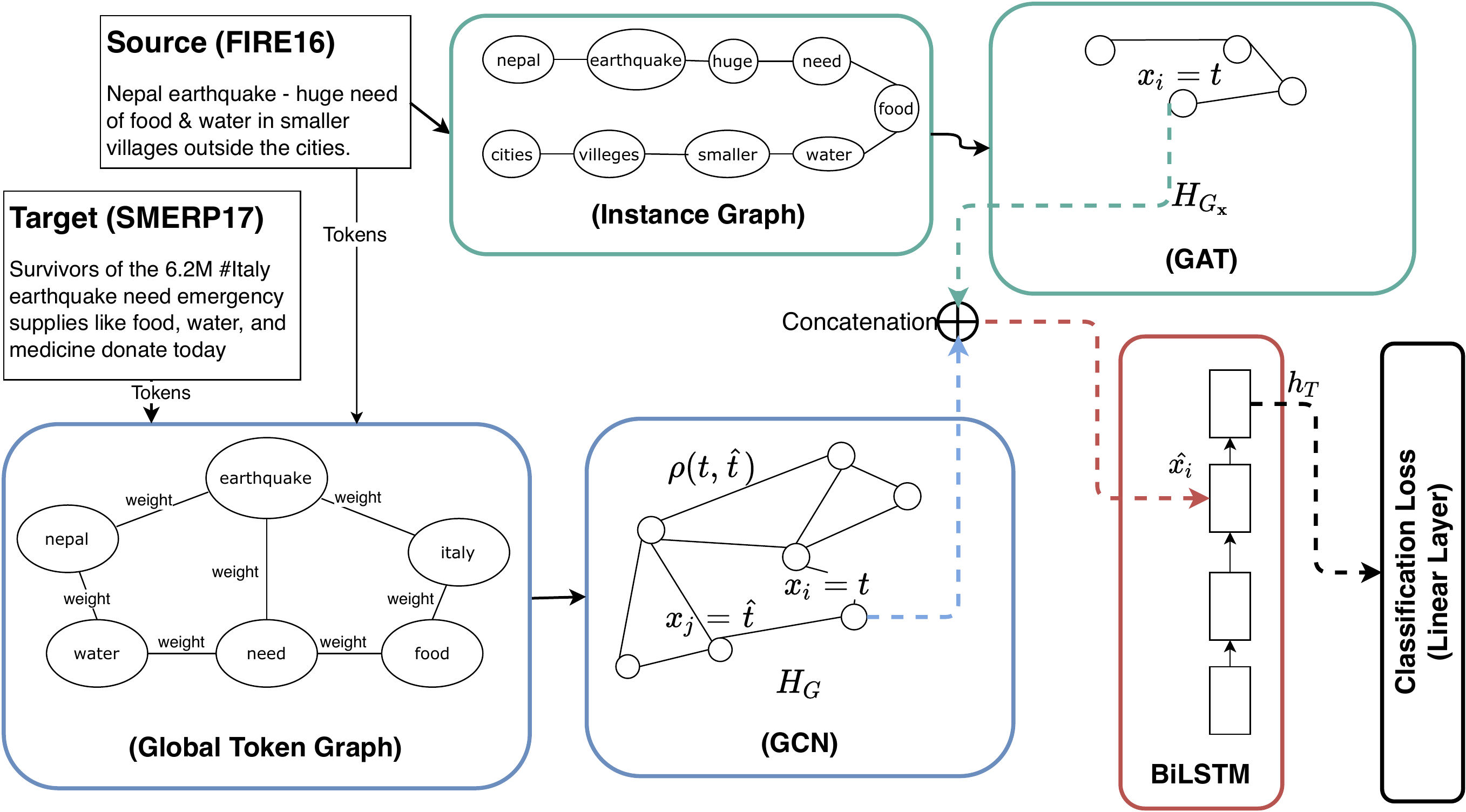}
    \caption{A block level illustration of our model. The components $H_G$ and $H_{G\mathbf{x}}$ represent the GCN on token graph (Section~\ref{subsec:global_token_graph}) and GAT on instance graph (Section~\ref{subsec:local_instance_graph}) respectively. Here $x_i$ is a token from instance $\mathbf{x}$. The BiLSTM layer (Section~\ref{subsec:classification_layer}) uses concatenation of the two to learn an end-to-end model. For simplicity, we only show connections in one direction and the last forward hidden state $h_T$.}
    \label{fig:model_arch}
\end{figure*}
\subsection{UDA for Disaster Response}
An additional challenge specific to disasters is the difficulty of obtaining labeled data in source domain. Annotating large amount of examples based on the type of actionable information is expensive.
Domain adaptive approaches in disaster domain should consider this constraint along with non-availability of labeled target domain data. As the amount of labeled data from source domain is limited, UDA approaches in disaster utilizes unlabeled source data. Whereas, majority of UDA approaches discussed in previous Subsection \ref{sec:related_uda} do not utilize source domain unlabeled data.
 
To classify tweets relevant to a disaster, \cite{qcri} utilized labeled and unlabeled data from source domain along with unlabeled data from target domain. In their proposed end-to-end approach, adversarial learning with gradient reversal and graph convolutional networks is utilized to achieve domain adaptation between a pair of source and target disaster. A document-level graph was constructed using k-nearest neighbours of Word2Vec \cite{word2vec} vectors. 
However, document level embedding might fail to capture the token-level distribution shift between source and target domain. SOTA performance in disaster scenario was achieved recently in \cite{sigir'20} where the authors use a domain reconstruction classification network (DRCN). DRCN reduces the covariate shift by reconstructing the target domain data with an LSTM based autoencoder. Reconstruction based approaches do not utilize token-level information. Although, several approaches in the literature explore domain adaptation setting for disaster relief, majority of disaster relief approaches were limited to predicting binary relevance of examples only.
To overcome these challenges in disaster domain, we propose a Graph Neural Network based approach which explicitly models both document-level local contexts to maximize class separation along with token-level global context to minimize the domain difference. We achieve this by using labeled source data for local context and unlabeled data from source and target domains for token-level global context. We apply our model on multiple datasets with varying amount of labeled data in source domain. Although availability of source domain labeled data is limited in disaster domain, this challenge has not been investigated before in disaster domain. We experiment on real world multi-label disaster tweet datasets where the amount of available labeled data in source domain is very limited (approx 1000 training examples or lower).
\section{Proposed Method}
We refer to the past disaster domain as source ($\mathbf{S}$) and current one as target ($\mathbf{T}$). We assume access to labeled data only in $\mathbf{S}$, in the form of $\{\mathbf{x}, y\}$, where $y$ is a label corresponding to the instance (e.g. tweet) $\mathbf{x}$. In this setting we would like to learn a classifier in domain $\mathbf{T}$, using labeled (and unlabeled) data from $\mathbf{S}$ and using only unlabeled data from $\mathbf{T}$. We refer to the labeled and unlabeled subsets of instances from $\mathbf{S}$ as $\mathbf{S}_L$ and $\mathbf{S}_U$ respectively and refer to unlabeled data from $\mathbf{T}$ as $\mathbf{T}_U$.  

Our model is composed of a two-part (\textit{Global} and \textit{Local}) graph based framework, trained in an end-to-end manner using the classification loss on labeled source data. Both these graphs are defined on the tokens available in the data. The \textit{Global}-part helps in learning domain agnostic feature representations of tokens in instances $\mathbf{x}$ using a large weighted graph constructed from tokens in $\mathbf{S}_L \cup \mathbf{S}_U \cup \mathbf{T}_U$. The edge weights in this graph are constructed based on token co-occurrence information. The \textit{Local}-part helps in preserving the local per-instance context in token representations using an unweighted graph constructed from the tokens of instance $\mathbf{x}$. 

One important point to consider here is that the global graph is weighted and the local graph is unweighted. This is because, the global graph is constructed from all the available posts from source and target domain. Since the number of such posts is large, the token co-occurrence information can be computed from a large sample, and can be indicative of the contextual similarity between the tokens. The same information can be further leveraged in the subsequent steps. 
On the other hand, the local graph is instance-specific and is constructed from individual posts. Co-occurrence of tokens in a single post (which is a short text) can be spurious, or also can happen by accident. Hence, although we retain the co-occurrence information in the local graph, we do not set any weight for the edges. However, we learn these weights using an attention framework whose parameters are learned during the training process.

We then combine representations of tokens obtained from these two graphs to form unified token representation that are both (a) domain agnostic and (b) retain per-instance level context. Figure \ref{fig:model_arch} illustrates an overview of our approach. Next, we describe the individual components of our model in detail.
\subsection{Global-Token Graph} \label{subsec:global_token_graph}
One of the key challenges in domain adaptation is to mitigate the covariate shift between domains. Formally, covariate shift is the difference between the feature conditional distributions in the two domains, i.e. $p_{\mathbf{S}}\left(\phi\left(\mathbf{x}\right) \mid y\right) \neq p_{\mathbf{T}}\left(\phi\left(\mathbf{x}\right) \mid y\right)$, where $\phi$ is an appropriate featurization of $\mathbf{x}$. As an example, consider the two tweets $\mathbf{x}_1=$``\texttt{I need water}'' and $\mathbf{x}_2=$``\texttt{I need wifi}'' from two different domains both associated with the label `\textit{Resources Required}'. The example illustrates the difference in kind of resources needed when a disaster (e.g. earthquake) hits two different locations with vastly different demographics and socio-economic status such as developing nation `Nepal' and developed world nation like `Italy'.

However, in unsupervised setting, we do not have access to labels in the target domain, and we address the covariate shift approximately by learning a domain agnostic feature representation $\phi$ of tokens $x_i$ across domains $\mathbf{S}$ and $\mathbf{T}$. As a concrete example, we would want $\phi(\textrm`\texttt{I need water}\textrm') \approx \phi(\textrm`\texttt{I need wifi}\textrm')$.

We form a weighted graph $G=(V,E,\mathrm{P})$. The vertex set $V$ consists of all the tokens from combined corpus in both domains i.e. $\mathbf{S}_L \cup \mathbf{S}_U \cup \mathbf{T}_U$. Two tokens $t$ and $\hat{t}$ are connected by an edge in $E$ if they co-occur within a context window of an instance in either of $\mathbf{S}_L,\mathbf{S}_U$ or $\mathbf{T}_U$. The weight of an edge $\rho\left(t,\hat{t}\right)$ is defined as, 
\begin{equation} \label{eq:weight_token_graph}
  \rho\left(t,\hat{t}\right) = \frac{1}{2} \left(\frac{n_{t,\hat{t}}^S}{n_t^S + n_{\hat{t}}^S} + \frac{n_{t,\hat{t}}^T}{n_t^T + n_{\hat{t}}^T} \right)  
\end{equation}

where, $n_{t,\hat{t}}^{S\mid T}$ are the co-occurrence counts of $t,\hat{t}$ within a context window in $\mathbf{S}$ and $\mathbf{T}$ respectively and $n_t$ (and $n_{\hat{t}}$) are the number of occurrences of the token $t$ (and $\hat{t}$) in the two domains. As these edge weights are estimated from a large corpus, we consider these weights to be reflective of underlying token co-occurrence frequency. The graph $G$ captures the notion of co-occurrence of tokens across two domains in its neighborhood. As a concrete example, from tweets $\mathbf{x}_1$ and $\mathbf{x}_2$, the token nodes $\texttt{water}$ and $\texttt{wifi}$ would be 2-hop neighbors of each other connected by the node $\texttt{need}$ in $G$. Also, the nodes $\texttt{nepal}$ and $\texttt{italy}$ are connected via $\texttt{earthquake}$ in Figure \ref{fig:model_arch}.

Taking advantage of this neighborhood in $G$, we define a Graph Convolutional Network (GCN) \cite{gcn} on $G$ to learn representations for the token nodes. In GCN, the node representations are learnt using the multi-hop neighborhood of a node. Therefore, GCN learns similar node representations in a neighborhood. Since the neighborhoods in $G$ are composed of tokens from $\mathbf{S}$ and $\mathbf{T}$, this results in domain agnostic feature representations for tokens in $\mathbf{x}$. Formally, we define a GCN \cite{gcn} as follows, 
\begin{align} \label{eq:gcn_global}
   H^{(l+1)} &= \sigma \left( D^{-\frac{1}{2}} \hat{A} D^{-\frac{1}{2}} H^{(l)}W^{(l)}\right)
\end{align}
with $H^{(0)} = \texttt{GloVe}(\mathbf{x})$. Here, $D$ is the diagonal degree matrix of $G$ and $\hat{A} = A + I$, where $A$ is the adjacency matrix of $G$. We restrict to $2$-hop neighborhood, thus $l=0,1$.

In our experiments, we will show qualitatively how the GCN based representations improve the GloVe \cite{glove} based representations by bringing the tokens, that appear in similar contexts in the two domains, closer to one another.
\subsection{Local-Example Graph} \label{subsec:local_instance_graph}
The global token graph layer discussed in Section~\ref{subsec:global_token_graph} builds domain agnostic feature representation of tokens in instances $\mathbf{x}$. However, it does not take into account the local context defined by tokens of $\mathbf{x}$ alone. The recent success of context-dependent word embedding such as BERT \cite{bert} in a wide range of NLP tasks indicate the importance of modelling the local context. 

To model the local context, we define an instance-level graph $G_\mathbf{x}=(V_\mathbf{x},E_\mathbf{x})$, where $V_\mathbf{x}$ is the set of tokens in an instance  $\mathbf{x}$ and are connected by an edge in the edge set $E_\mathbf{x}$ if they appear within a context window of each other. It is noteworthy that $G_\mathbf{x}$ is an unweighted graph because it would be challenging to apply a weighting scheme 
owing to the local context.
Instead, we propose to learn the weights in the form of attention scores on the edges in a Graph Attention Network (GAT) \cite{gat} framework. The GAT on the local instance graphs $G_{\mathbf{x}}$ learns node representations for tokens in instance $\mathbf{x}$ along with the edge attention scores in an end-to-end manner. Formally, we define a $k$-layer GAT on $G_{\mathbf{x}}$ parameterized by $W^k$ as follows,
\begin{align} \label{eq:gat_local}
    h^{(l+1)}_i &= \bigoplus_{k} \sum_{j\in N(i)} \alpha^{k}_{i,j} W^{k} h^{(l)}_{j}
\end{align}
with $h^{0}_i = \texttt{GloVe}(x_i)$. Here, $h^{(l)}_i$ is the $l^{th}$-layer representation of token node $x_i$, $N(i)$ are tokens belonging to context window around the token $x_i$ and $\oplus$ indicates concatenation. We restrict to a 2-layer GAT i.e. $l={0,1}$ and use two attention heads in first layer and a single head in the second layer. We consider the context window around token $x_i$ as $[x_{i-1}\ldots x_{i+1}]$. Figure \ref{fig:model_arch} shows such an instance graph with window size $2$ for a tweet posted during 2015 Nepal earthquake.

It is worthwhile to note that GAT is closely related to the self-attention mechanism in Transformer based architectures \cite{transformer}. The key advantage of GAT over transformers in our setting is the fact that GAT's attention mechanism is restricted to the neighborhood of a token, which in our case is the context window $[x_{i-1}\ldots x_{i+1}]$. In comparison, transformer based models learn attention scores over the entire sequence at every position in the sequence and require much larger number of parameters.
\subsection{Classification Layer} \label{subsec:classification_layer}

In Section~\ref{subsec:global_token_graph} and Section~\ref{subsec:local_instance_graph}, we focused on defining domain agnostic representations of tokens $x_i$ and on preserving the local instance context respectively. We finally combine these two representations by concatenating them as follows: 
\begin{equation}
     h(x_i) = h^{(2)}_{G_\mathbf{x}}(i) \oplus h^{(2)}_{G}(j)
\end{equation}
where, $j$ is the index of the token $x_i$ in the token graph $G$. 
Also, $h^{(2)}_{G_\mathbf{x}}$ and $h^{(2)}_{G}$ are final layer representations from local instance graph GAT (Eq. ~\eqref{eq:gat_local}) and global token graph GCN (Eq. ~\eqref{eq:gcn_global}) respectively.  
The concatenated representation combines the global domain agnostic and the local context dependent nature of the tokens in instances $\mathbf{x}$. 

Using these representations $h(x_i)$ as input features in a single layer bi-directional LSTM (BiLSTM) network, we learn an end-to-end model. The parameters of the global token graph (the underlying GCN) and local instance graph (the underlying GAT) are updated in the backward pass using standard cross-entropy loss on the instances from source domain $\mathbf{S}$. We use the last hidden state of the BiLSTM as the representation of the instance $\mathbf{x}$ and pass it through a single fully connected layer to predict the class labels. 
%
\section{Experiments} \label{sec:experiment_details}
The main empirical focus of our work is to measure the effectiveness of our approach when no labeled data is available from target domain. To this effect, considering an UDA setting, we pair up public datasets from disaster management as source ($\mathbf{S}$) and target ($\mathbf{T}$) for our experiments and supervise our model (refer to Section~\ref{subsec:classification_layer}) on only labeled instances from $\mathbf{S}$ and test on $\mathbf{T}$. While previous works are restricted to datasets involving only binary classification \cite{qcri, sigir'20}, we also report results on public datasets involving more realistic and useful multi-label multi-class classification setting. Additionally, to verify the effectiveness of our approach in limited labeled data scenarios we reduce the source domain training data. We consider $50\%$, $25\%$ and $10\%$ of original training set without changing the validation and test sets. 
%
\subsection{Datasets} \label{subsec:datasets}
\begin{table}[t]
\centering
\begin{adjustbox}{totalheight=2.\height
}
    \begin{tabular}{@{}lrr|rrr@{}}
    \toprule
     \multicolumn{1}{l|}{\textbf{Dataset}}  & \textbf{1} & \textbf{0} & \textbf{Train} & \textbf{Val} & \textbf{Test} \\ \cmidrule(l){2-6} 
    \textbf{NEQ} & 5527 & 6141 & 7000 & 1166 & 3502 \\
    \textbf{QFL} & 5414 & 4619 & 6019 & 1003 & 3011 \\ \bottomrule
    \end{tabular}
    \end{adjustbox}
\caption{Stats of NEQ and QFL datasets. The class $1$ indicates tweets relevant to disaster and $0$ indicates otherwise.}
\label{table:neq_qfl_details}
\end{table}
On datasets involving binary classification, we conduct experiments on two publicly available datasets of tweets; `2015 Nepal earthquake' (NEQ) and `2013 Queensland flood' (QFL) \cite{qcri}. The associated class information indicates if a tweet is relevant to the disaster, and the task is to predict the (ir)relevance label for tweets in the test set. Table~\ref{table:neq_qfl_details} presents the details of NEQ and QFL datasets. The datasets also include unlabeled tweets, mined in the same time-frame of the disaster, in the form of tweet ids (for exact details, refer to~\cite{qcri}). We used Twitter's public API to download these tweets. The process resulted in $49,223$ and $15,464$ unlabeled tweets related to NEQ and QFL datasets, respectively. We used the same train, dev and test split as provided by the authors for our experiments.

%
\begin{table}[t]
\centering
\begin{adjustbox}{totalheight=1.6\height
}
\begin{tabular}{@{}lccl@{}}
\toprule
\multicolumn{2}{c|}{\textbf{FIRE16}}        & \multicolumn{2}{c}{\textbf{SMERP17}}                      \\ \midrule
\textbf{Class name}          & \textbf{ID} & \textbf{ID}        & \textbf{Class name}                  \\ \midrule
Resources Available          & 1           & \multirow{2}{*}{1} & \multirow{2}{*}{Resources Available} \\ 
Medical Resources Available  & 3           &                    &                                      \\ \midrule
Resources Required           & 2           & \multirow{2}{*}{2} & \multirow{2}{*}{Resources Required}  \\ 
Medical Resources Required   & 4           &                    &                                      \\ \midrule
Resources Specific Locations & 5           & -                  &                                      \\ \midrule
Infrastructure Damage \& Restoration & 7 & 3 & Infrastructure Damage \& Restoration                    \\ \midrule
Activities NGOs / Government & 6           & 4                  & Rescue Activities NGOs / Government \\ \bottomrule
\end{tabular}
\end{adjustbox}
\caption{Class mapping from FIRE16 to SMERP17. Class $5$ of FIRE16 was ignored.}
\label{table:smerp17_fire16_map}
\end{table}
On datasets involving multi-class multi-label classification, we conduct experiments on `Forum for Information Retrieval Evaluation 2016' (FIRE16)~\cite{fire16} and `Social Media for Emergency Relief and Preparedness' (SMERP17)~\cite{smerp17} containing tweets posted during 2015 earthquake in Nepal and 2016 earthquake in central Italy respectively. These datasets contain tweets with granular, actionable class information such as resource available, resource required, infrastructure damage, and rescue activities. FIRE16 contains $7$ granular class information, whereas SMERP17 contains $4$ classes that are closely related to the FIRE16 classes. Due to lack of publicly available datasets with granular actionable information, we pair up FIRE16 and SMERP17 by mapping the $7$ classes of FIRE16 into $4$ classes of SMERP17. The grouping is quite natural, and the details are presented in Table~\ref{table:smerp17_fire16_map}. Once again, these datasets also include unlabeled tweets, and we followed a similar process as for binary datasets to collect tweets. This process resulted in $68,964$ unlabeled tweets for SMERP17. Details of the FIRE16 and SMERP17 datasets after mapping are provided in Table \ref{table:fire16_details} and  \ref{table:smerp17_details}. Note that the example counts might differ from original datasets. As Twitter policy does not allow direct sharing, tweets were downloaded before the experiments, and some tweets may not be retrieved if they are deleted or made private.
\begin{table}[!htbp]
    \centering                          
    \begin{adjustbox}{totalheight=2.0\height, width=0.8\linewidth}
    \begin{tabular}{@{}lrrrrr@{}}
        \toprule
        \textbf{FIRE16} & \textbf{Train} & \textbf{Val} &\textbf{Test} &\textbf{Class Dist(\%)} \\ \midrule
        1 & 498 & 55  & 237 & 35.94 \\
        2 & 217 & 24  & 104 & 15.69 \\
        3 & 367 & 41  & 175 & 26.52 \\
        4 & 302 & 34  & 144 & 21.83 \\ \midrule
        \# Examples & 957 & 106 & 459 & \\ \bottomrule
   
        \end{tabular}
        \end{adjustbox}
    \caption{Details of FIRE16 dataset.}
    \label{table:fire16_details}
\end{table}

\begin{table}[!htbp]
    \centering                          
    \begin{adjustbox}{totalheight=2.0\height, width=0.8\linewidth}
    \begin{tabular}{@{}lrrrrr@{}}
        \toprule
        \textbf{SMERP17} & \textbf{Train} & \textbf{Val} &\textbf{Test} &\textbf{Class Dist(\%)} \\ \midrule
        1 & 184 & 22  & 76  & 13.60 \\
        2 & 105 & 15  & 46  & 8.00  \\
        3 & 774 & 141 & 393 & 63.09 \\
        4 & 212 & 25  & 80  & 15.29 \\ \midrule
        \# Examples & 1159 & 189 & 548 & \\ \bottomrule
   
        \end{tabular}
        \end{adjustbox}
    \caption{Details of SMERP17 dataset.}
    \label{table:smerp17_details}
\end{table}

\begin{table*}[!htbp]
\centering
\begin{adjustbox}{totalheight=2.3\height
}
\begin{tabular}{@{}ccrrrrrrrrr@{}}
\toprule
 &  & \multicolumn{4}{c|}{\textbf{F$_{1}$ Weighted}} & \multicolumn{2}{c|}{\textbf{F$_{1}$ Micro}} & \multicolumn{2}{c}{\textbf{F$_{1}$ Macro}} \\ \cmidrule(l){3-6} \cmidrule(l){7-8} \cmidrule(l){9-10}
\textbf{Source} & \textbf{Target} & \textbf{DAAT} & \textbf{DRCN} & \textbf{BERT} & \textbf{GLEN} & \textbf{BERT} & \textbf{GLEN} & \textbf{BERT} & \textbf{GLEN} \\ \midrule
\textbf{NEQ} & \textbf{QFL} & 65.90 & \underline{81.18} & 80.72 & \textbf{83.42} & 80.71 & \textbf{83.52} & 80.72 & \textbf{83.48}\\
\textbf{QFL} & \textbf{NEQ} & 59.50 & \underline{68.38} & 67.22 & \textbf{71.61} & 67.80 & \textbf{71.85} & 67.25 & \textbf{71.62}\\ \cmidrule(r){1-2}
\textbf{NEQ} & \textbf{NEQ} & 65.11 & - & 76.39 & \textbf{77.76} & 76.37 & \textbf{77.78} & 76.33 & \textbf{77.68}\\
\textbf{QFL} & \textbf{QFL} & 93.54 & - & 96.24 & \textbf{96.77} & 96.24 & \textbf{96.77} & 96.22 & \textbf{96.75}\\ \bottomrule
\end{tabular}
\end{adjustbox}
\caption{Weighted, Micro and Macro F$_{1}$ scores for NEQ and QFL datasets. 
Proposed method GLEN outperforms other SOTA methods in both cross-domain and in-domain settings for both datasets.}
\label{table:neq_qfl_f}
\end{table*}
\begin{table*}[!htbp]
\centering
\begin{adjustbox}{totalheight=2.3\height
}
\begin{tabular}{@{}ccrrrrrrrr@{}}
\toprule
&  & \multicolumn{2}{c|}{\textbf{F$_{1}$ Weighted}} & \multicolumn{2}{c|}{\textbf{F$_{1}$ Micro}} &  \multicolumn{2}{c}{\textbf{F$_{1}$ Macro}} \\ \cmidrule(l){3-4} \cmidrule(l){5-6} \cmidrule(l){7-8}
\textbf{Source} & \textbf{Target}& \textbf{BERT} & \textbf{GLEN} & \textbf{BERT} & \textbf{GLEN} & \textbf{BERT} & \textbf{GLEN} \\ \midrule
\textbf{FIRE16} & \textbf{SMERP17} & 76.21 & \textbf{80.80} & 76.47 & \textbf{76.30} & 52.00 & \textbf{55.93} \\
\textbf{SMERP17} & \textbf{FIRE16} & 55.52 & \textbf{56.56} & 56.89 & \textbf{58.43} & 54.13 & \textbf{55.07} \\ \cmidrule(r){1-2}
\textbf{FIRE16} & \textbf{FIRE16} & 77.36 & \textbf{82.04} & 77.62 & \textbf{81.92} & 76.14 & \textbf{80.83} \\
\textbf{SMERP17} & \textbf{SMERP17} & 91.68 & \textbf{93.37} & 91.93 & \textbf{94.21} & 82.04 & \textbf{86.46} \\ \bottomrule
\end{tabular}
\end{adjustbox}
\caption{Weighted, Micro and Macro F$_{1}$ scores for FIRE16 and SMERP17 datasets. Proposed method GLEN outperforms BERT in all possible settings.}
\label{table:fire16_smerp17_f}
\end{table*}
\subsection{Baselines \& Models} 
\label{subsec:baseline}
We compare our approach against SOTA methods in disaster domain. For binary classification, we compare against the methods named DRCN and DAAT presented in \cite{sigir'20} and \cite{qcri} respectively. These methods, to the best of our knowledge, are the most recent methods with SOTA performance on NEQ and QFL datasets. In addition, we also compare our work against BERT \cite{bert}. As BERT outperforms other deep learning based methods (e.g. RNN, CNN) on NEQ-QFL dataset pair, shown by authors Li et al. of DRCN \cite{sigir'20} in \cite{bert_compare} w.r.t. F$_1$ score, we did not run experiments on these method. On multi-label datasets, owing to lack of any baseline, we compare only against BERT. To the best of our knowledge, we are the first to compare against BERT in disaster domain literature on multi-label domain adaptation setting. We fine-tune all layers of the BERT-base-uncased model using the source domain labeled data by appending a linear layer as classification head to the BERT model.

To understand the efficacy of various modelling components of our approach, we report experimental results on the following ablations (all of which include the classification layer with cross-entropy loss over the labelled data as in Section~\ref{subsec:classification_layer}). The last one being the overall model:
\begin{itemize}[leftmargin=*]
    \setlength\itemsep{0em}
    \item \textbf{LEN}: Only Local-Example (Section~\ref{subsec:local_instance_graph}) Graph component, without the Global-Token Graph i.e. trained on $\mathbf{S}_L$ using GAT only.
    \item \textbf{S-GLEN}: The Global-Token Graph constructed from tokens in $\mathbf{S}$ alone (i.e. $\mathbf{S}_L \cup \mathbf{S}_U$), along with Local-Example Graph component. GCN over $\mathbf{S}_L \cup \mathbf{S}_U$ with GAT over $\mathbf{S}_L$.
    \item \textbf{T-GLEN}: The Global-Token Graph constructed from tokens in $\mathbf{S}_L \cup \mathbf{T}_U$, along with Local-Example Graph component. GCN over $\mathbf{S}_L \cup \mathbf{T}_U$ with GAT over $\mathbf{S}_L$.
    \item \textbf{GLEN}: The Global-Token Graph constructed from tokens in  $\mathbf{S}_L \cup \mathbf{S}_U \cup \mathbf{T}_U$ along with Local-Example Graph component, i.e. the overall proposed model. GCN over $\mathbf{S}_L \cup \mathbf{S}_U \cup \mathbf{T}_U$ with GAT over $\mathbf{S}_L$.
\end{itemize}
%
\subsection{Training Configuration} Hyperparameters such as learning rate and the number of epochs were tuned with the help of validation data with patience parameter set to $4$. As we assumed no labeled data was available from target domain, we use the source domain validation set for tuning. labeled data from target domain was used during testing only. Learning rate search space was limited to $\{10^{-3}$, $10^{-4}$, $10^{-5}\}$ values only.
We trained BERT on a system with a `Nvidia Tesla P100' GPU with $12$GB GPU-RAM, $96$GB system RAM and $56$ cores. Our model GLEN was trained on CPU only with $32$GB RAM and $20$ cores. We used $100$ as hidden dimension size for all experiments and $300$ dimensional GloVe embeddings were used as initial input token-level features. The number of parameters for GLEN was $\approx664$k which is $\approx165$ times smaller than BERT in terms of number of parameters. All results reported for GLEN and BERT are averaged across $3$-runs, each corresponding to a different random seed for the experiment while training. Below we present the architecture details of our model GLEN:
\small{
\begin{verbatim}
GLEN(
  (token_gcn): GCN(
    (gc1): GraphConvolutionLayer (300 -> 100)
    (gc2): GraphConvolutionLayer (100 -> 100)
  )
  (instance_gat_dgl): Instance_GAT_dgl(
    (conv1): GATConv(
      (fc): Linear(in_features=300,
                   out_features=200, bias=False)
      (feat_drop): Dropout(p=0.0, inplace=False)
      (attn_drop): Dropout(p=0.0, inplace=False)
      (leaky_relu): LeakyReLU(neg_slope=0.2)
    )
    (conv2): GATConv(
      (fc): Linear(in_features=200,
                   out_features=100, bias=False)
      (feat_drop): Dropout(p=0., inplace=False)
      (attn_drop): Dropout(p=0., inplace=False)
      (leaky_relu): LeakyReLU(neg_slope=0.2)
    )
  )
  (bilstm_classifier): BiLSTM_Classifier(
    (lstm): LSTM(200, 100, num_layers=2, 
                 batch_first=True, dropout=0.2,
                 bidirectional=True)
    (fc): Linear(in_features=200,
                 out_features=4, bias=True)
  )
)
\end{verbatim}
}
%
\begin{figure*}
\subfloat[Before training]{\includegraphics[width = 3.5in]{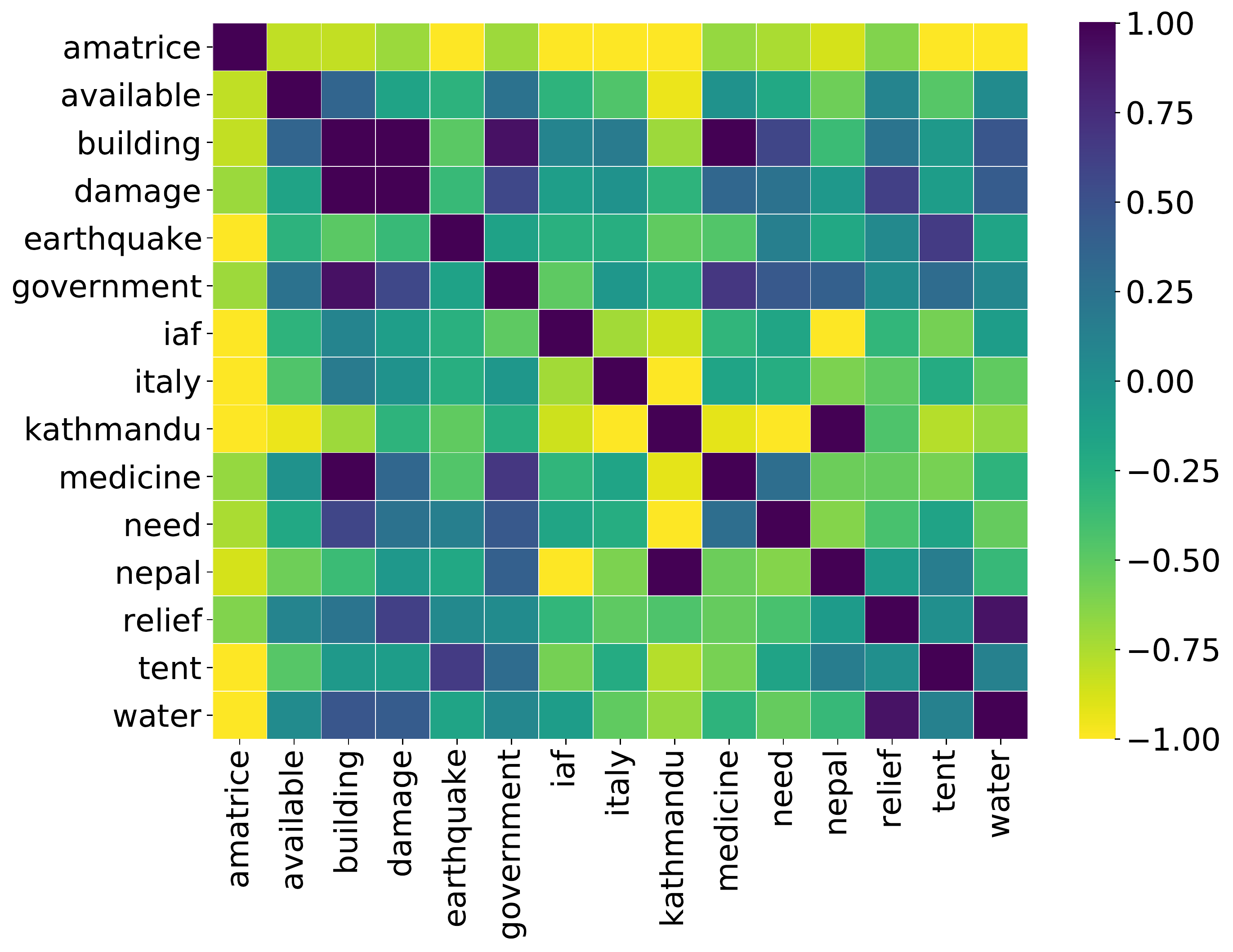}} 
\subfloat[After training]{\includegraphics[width = 3.5in]{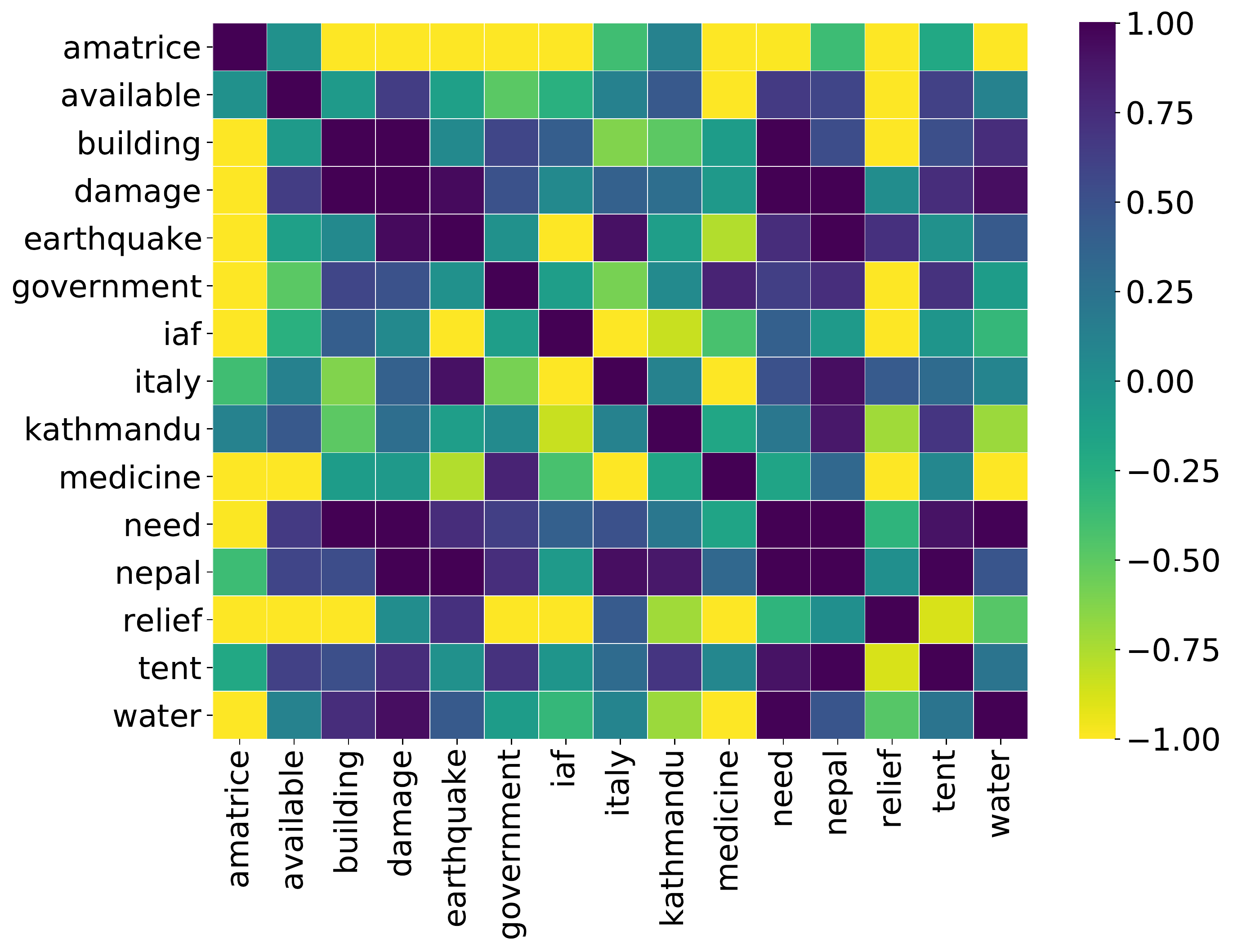}}\\
\caption{Heatmap of normalized cosine similarity of a sample of token vocabulary from FIRE16 and SMERP17 dataset pair; (a) before and (b) after global token graph training. We observe that the training results in tokens forming clusters of similarity (indicated by clusters of \textit{warmer} values on the right). The normalization scheme is discussed in Section~\ref{subsec:qual_results}.}
\label{fig:heatmap}
\end{figure*}
\subsection{Results} \label{subsec:results}
We report results on three different experiments. Initially we compare the performance of our model with baselines when all source domain training data was available. Then we verify the effectiveness of our approach in low-labeled data regime, i.e. limited training data. We also performed ablation study over different global embedding settings. Finally, we performed qualitative investigation on the capabilities of our model.
\begin{table*}[hbt!]
\centering
\begin{adjustbox}{totalheight=2.3\height
}
\begin{tabular}{@{}llrrrrrrrrrrr@{}}
\toprule
&  & \multicolumn{3}{c|}{\textbf{$50\%$}} & \multicolumn{3}{c|}{\textbf{$25\%$}} & \multicolumn{3}{c}{\textbf{$10\%$}} \\ \cmidrule(l){3-5} \cmidrule(l){6-8} \cmidrule(l){9-11}
\textbf{Source} & \textbf{Target}& \textbf{BERT} & \textbf{GLEN} & \textbf{Gain \%} & \textbf{BERT} & \textbf{GLEN} & \textbf{Gain \%} & \textbf{BERT} & \textbf{GLEN} & \textbf{Gain \%} \\ \midrule
\textbf{NEQ} & \textbf{QFL} & 80.49 & \textbf{83.36} & 2.87 & 79.45 & \textbf{82.52} & 3.07 & 78.03 & \textbf{81.86} & 3.82 \\
\textbf{QFL} & \textbf{NEQ} & 65.78 & \textbf{70.33} & 4.55 & 65.30 & \textbf{69.86} & 4.56 & 64.62 & \textbf{69.40} & 4.78 \\
\textbf{NEQ} & \textbf{NEQ} & 74.07 & \textbf{75.82} & 1.75 & 71.77 & \textbf{75.01} & 3.24 & 71.36 & \textbf{74.19} & 2.83 \\
\textbf{QFL} & \textbf{QFL} & 94.73 & \textbf{96.23} & 1.59 & 94.53 & \textbf{96.14} & 1.60 & 94.09 & \textbf{95.72} & 1.63 \\ \cmidrule(r){1-2}
\textbf{FIRE16} & \textbf{SMERP17} & 72.64 & \textbf{76.77} & 4.13 & 63.05 & \textbf{74.67} & 11.61 & 27.42 & \textbf{58.23} & 30.81 \\
\textbf{SMERP17} & \textbf{FIRE16} & 37.73 & \textbf{47.33} & 9.59 & 28.21 & \textbf{43.24} & 15.03 & 16.71 & \textbf{33.81} & 17.10 \\
\textbf{FIRE16} & \textbf{FIRE16} & 70.44 & \textbf{78.68} & 8.23 & 58.99 & \textbf{72.25} & 13.26 & 42.73 & \textbf{57.08} & 14.35 \\
\textbf{SMERP17} & \textbf{SMERP17} & 91.24 & \textbf{93.49} & 2.24 & 84.03 & \textbf{88.02} & 3.99 & 71.29 & \textbf{77.14} & 5.84 \\ 
\midrule
Average &&&& 4.37 &&& 7.05 &&& 10.15 \\
\bottomrule
\end{tabular}
\end{adjustbox}
\caption{Weighted F$_{1}$ scores for NEQ-QFL and FIRE16-SMERP17 dataset pairs for same and cross domain setting in limited training data regimes. It can be observed that performance gap between our method GLEN and BERT increases with decrease in training data. Average row signifies average performance gain on all four datasets combined.}
\label{table:fire16_smerp17_limited}
\end{table*}
\subsubsection{Results on all available data}
Table~\ref{table:neq_qfl_f} presents a comparison of GLEN with other SOTA methods (DAAT and DRCN) and BERT on the test set of the NEQ and QFL datasets. We observe that for cross-domain setting of NEQ-QFL dataset pair, GLEN outperforms both SOTA methods and BERT by $2.74\%$ and $3.55\%$ on average, respectively, w.r.t. weighted F$_{1}$ metric. In addition, we observe that GLEN outperforms other methods in same-domain (last two rows of Table~\ref{table:neq_qfl_f}) by $0.95\%$. The larger gain in cross-domain compared to same-domain setting indicates that our approach is better suited to domain adaptive setting, which is our original objective.
Table~\ref{table:fire16_smerp17_f} presents a comparison of GLEN with BERT on the test set of FIRE16 and SMERP17 dataset pair. Since we are the first to investigate UDA in the cross-domain multi-label setting, there are no baselines available and we compare against BERT. We observe that GLEN outperforms BERT in all settings on average $3.00\%$ w.r.t. weighted F$_{1}$ metric. Once again, GLEN is better than BERT when we train and test on the same domain, showing the robust performance of the proposed approach in general.
\subsubsection{Results on limited training data}
\begin{figure*}
\subfloat[NEQ-QFL]{\includegraphics[width = 3.5in]{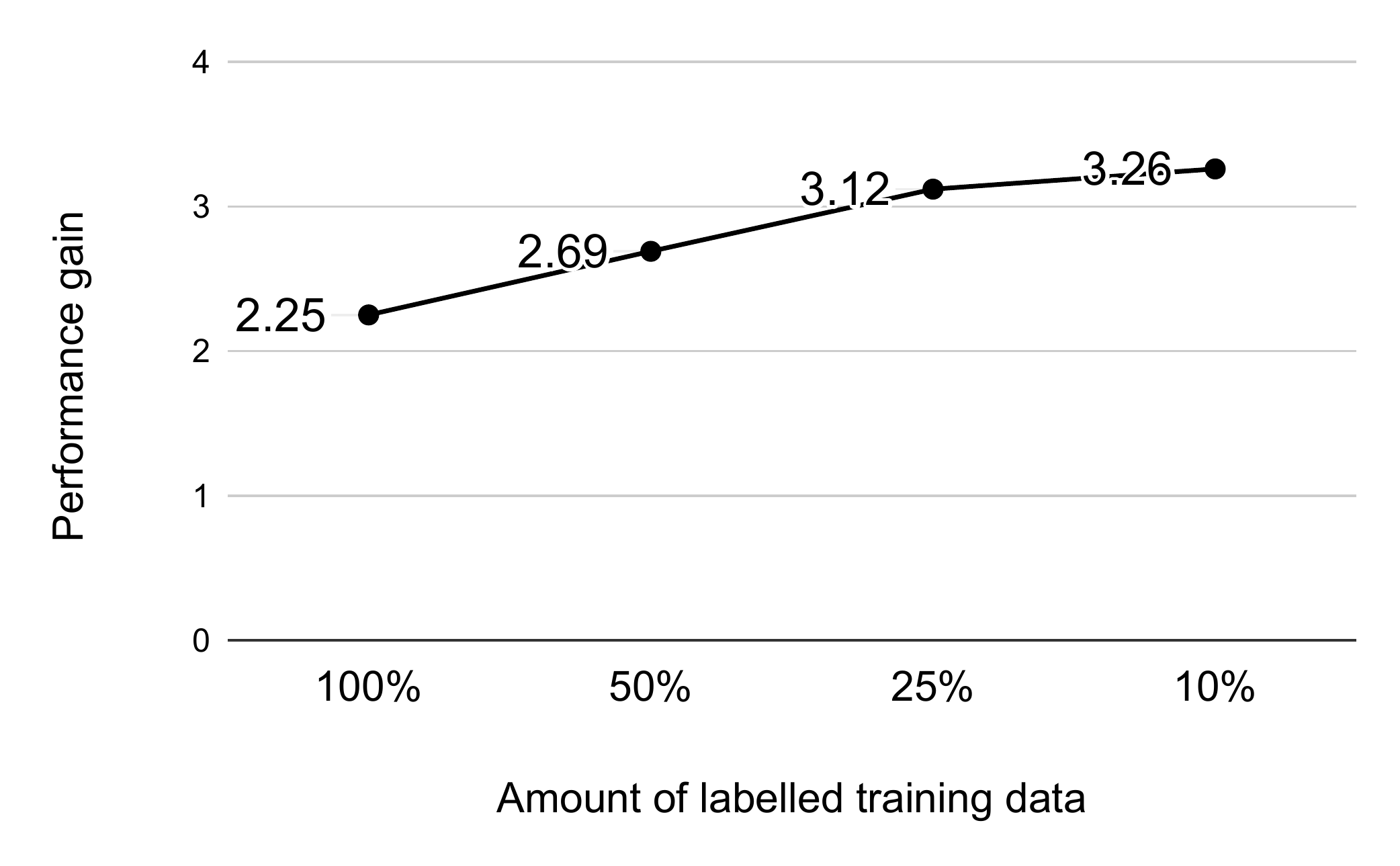}} 
\subfloat[FIRE16-SMERP17]{\includegraphics[width = 3.5in]{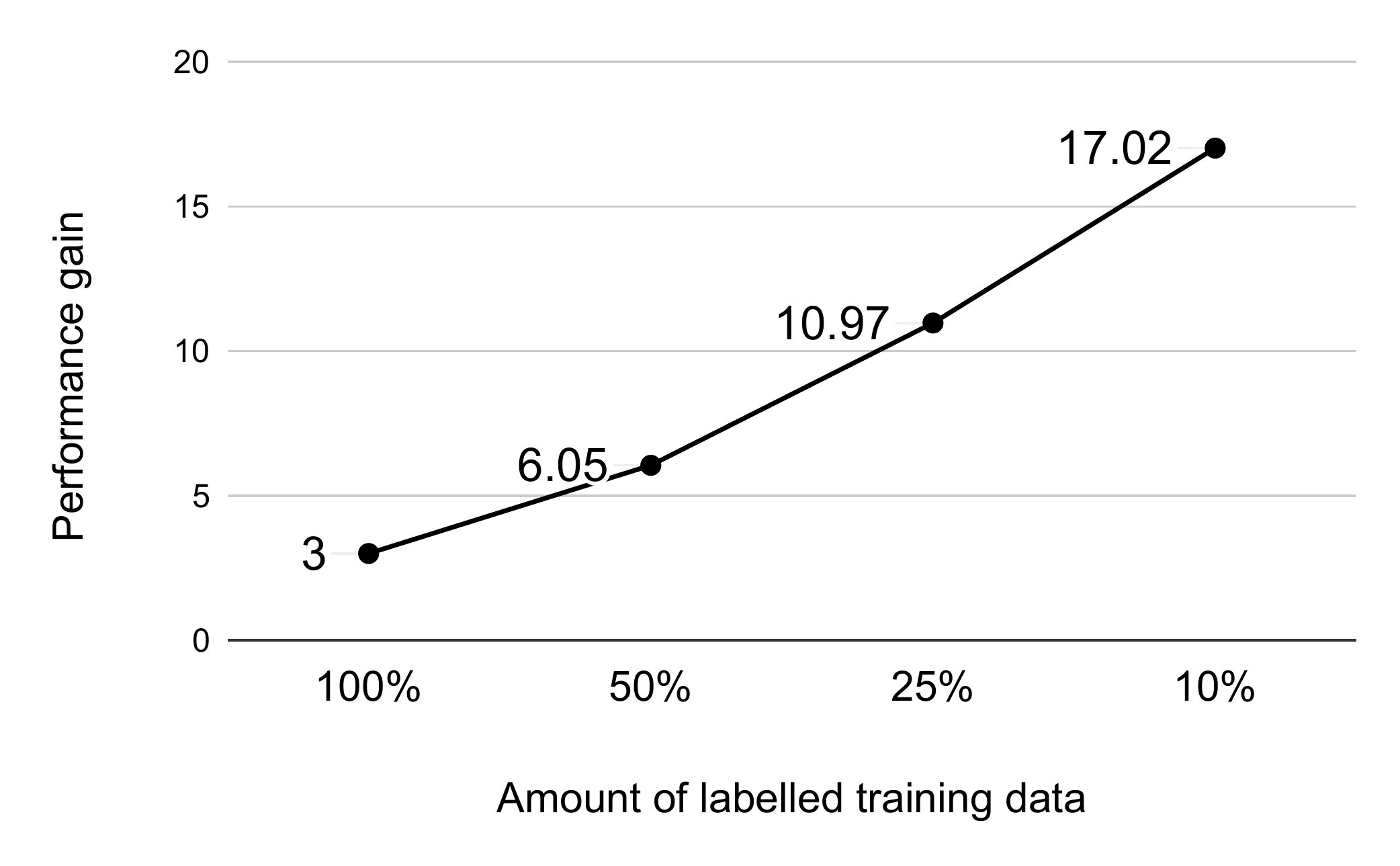}}\\
\caption{Average absolute performance gain (weighted F$_1$) of GLEN over BERT as training data is reduced on NEQ-QFL (left) and FIRE16-SMERP17 (right) dataset pairs respectively.}
\label{fig:perf_gain}
\end{figure*}

Table~\ref{table:fire16_smerp17_limited} summarises the performance of our approach compared to BERT when limited labeled data is available. We restrict the source domain training data to $100\%$, $50\%$, $25\%$ and $10\%$ of original training data size. We observed that as the amount of labeled training data decrease in source domain, the performance gain of GLEN over BERT increases as presented in Figure \ref{fig:perf_gain}. When the models are trained with $10\%$ training data i.e. $\approx 650$ examples for NEQ-QFL and $\approx 100$ examples for FIRE16-SMERP17 dataset pairs, the performance gain is as much as $3.26\%$ and $17.02\%$ respectively. The results indicate that GLEN can retain performance by utilizing the global token graph effectively. Between binary (NEQ, QFL) and multi-class (FIRE16, SMERP17) setups, the required annotation effort is more for the multi-label multi-class setup. Moreover, with the same effort, the number of examples available per-class is lesser in case of multi-label annotation. We see that the performance difference between GLEN and BERT is more in such scenarios. These findings indicate that GLEN is better in learning from smaller amount of labeled data, which is crucial for disaster management. 
%
\subsubsection{Component-wise results}
\begin{table}[t]
\centering
\begin{adjustbox}{totalheight=2.2\height, width=0.99\linewidth}
\begin{tabular}{@{}ccrrrrrr@{}}
\toprule
\textbf{Source} & \textbf{Target}  & \textbf{LEN}       & \textbf{S-GLEN}     & \textbf{T-GLEN}     & \textbf{GLEN} \\ \midrule
\textbf{NEQ} & \textbf{QFL} & 81.94  & \underline{82.58} & 81.97 & \textbf{83.42} \\
\textbf{QFL} & \textbf{NEQ} & 64.75  & \underline{65.44} & 65.13 & \textbf{71.61} \\
\cmidrule(r){1-2}
\textbf{FIRE16} & \textbf{SMERP17} & 74.04 & \underline{75.02} & 74.33 & \textbf{80.80} \\
\textbf{SMERP17} & \textbf{FIRE16} & 52.41 & \underline{56.15} & 55.83 & \textbf{56.56} \\
\bottomrule
\end{tabular}
\end{adjustbox}
\caption{Ablation with weighted F$_{1}$ scores for various cross-domain settings. Every GLEN model configuration is better than LEN, S-GLEN and T-GLEN.}
\label{table:all_ablation}
\end{table}
In Table~\ref{table:all_ablation}, we present an ablation over various components of the proposed model GLEN. We note that the introduction of global-token graph improves the performance of the model significantly as every model configuration performs better than LEN, which only uses instance (e.g. tweet) level information. In the model configurations, which include the global-token graph, GLEN is better than both S-GLEN and T-GLEN indicating that constructing a graph with tokens from both domains and propagating information is helpful. Performance of T-GLEN is worse in comparison to S-GLEN for every dataset combination; this is possibly due to low vocabulary overlap between labeled source and unlabeled target tokens. Resulting in sparser graph and hence the global-token graph representations can not capture much additional information over original Glove embeddings.
\subsection{Qualitative Investigation} \label{subsec:qual_results}
In Section~\ref{subsec:global_token_graph}, we claimed that GCN on the global-token graph should bring tokens which appear in similar contexts closer to one other in embedding space. We argued that this is beneficial to mitigate the covariate shift between the domains. We investigate if this holds in practice by designing an experimental study.
We selected a small subset of tokens from vocabulary and collected their representation before and after training the network.
The token subset is chosen on the basis of a heuristic where some token pairs appear in similar contexts from instances in the datasets. In contrast, some others do not appear in similar contexts. The particular heuristic helps investigate if the GCN component of the model is able to learn domain agnostic representation for tokens. We then calculated the cosine similarity of a token to all the other tokens in the subset using the input embedding (in case of "before") and output of global token graph (GCN outputs in case of "after"). The cosine similarity values are z-score normalized to have $0$-mean and $1$-std. deviation. The normalization was performed to investigate the relative similarity.

Figure~\ref{fig:heatmap} shows the heatmap of the normalized cosine similarity between pairs of tokens in the selected token subset. Figure~\ref{fig:heatmap}a is cosine similarity between the initial input (Glove) representation of a pair of tokens. At the same time, Figure~\ref{fig:heatmap}b shows the same for the token representations from the global-token graph (GCN component) after end-to-end training of the model. We note that the global-token graph representations form clusters of high and low similarities (indicated by values near $+1$ and $-1$ respectively), while input similarities are more uniformly spread out for each token indicating very similar values of cosine similarities. The similarities are also higher for token pairs which appear in similar contexts across instances from two datasets and are low for tokens which do not. 
Few concrete examples which illustrate this point better are shown in Table~\ref{table:cosine_sim_compare}. For example, the tokens \texttt{nepal} and \texttt{amartice} has normalized similarity score of $-0.9593$ before the training. However, as these two tokens generally appear in similar contexts (i.e. \texttt{earthquake}) they come closer after the training with normalized similarity score $-0.1583$. Similar observations can be seen for other tokens in Table~\ref{table:cosine_sim_compare}. This shows that our approach can bring tokens from different domains, appearing in similar context, closer. Interestingly, we also observe the opposite effects in case of the token pair \texttt{need} and \texttt{available}. As FIRE16-SMERP17 dataset pair have classes representing 'resources need' and 'resources available', our model decreased similarity to achieve class separation during training which is desirable in this scenario.
%
%
\begin{table}[!ht]
\centering
\begin{adjustbox}{totalheight=2.2\height, width=0.9\linewidth}
\begin{tabular}{@{}lrrr@{}}
\toprule
\textbf{Token Pair} & \textbf{Type} & \textbf{\shortstack{Glove \\ Similarity}} & \textbf{\shortstack{GLEN \\ Similarity}}\\ \midrule
(nepal, amatrice) & \textbf{(+)ve} & -0.9593 & -0.1583 \\
(tent, medicine) & \textbf{(+)ve} & -0.4485 & 0.2096 \\
(need, available) & \textbf{(-)ve} & 0.9934 & 0.6518 \\
(medicine, building) & \textbf{(-)ve} & -0.2639 & -0.3050 \\ \bottomrule
\end{tabular}
\end{adjustbox}
\caption{Normalized cosine similarity values for token pairs from Figure~\ref{fig:heatmap} for Glove and GLEN's Global-Token Graph representations. The \textit{Type} column indicates whether the tokens should come closer  (or move away) in embedding space, basis how they appear in common context.}
\label{table:cosine_sim_compare}
\end{table}
%
\section{Conclusion}
In this paper, we addressed the problem of unsupervised domain adaptation in the area of disaster management on online social networks. To this effect, we proposed a novel two-part graph-based framework which utilizes both labeled and unlabeled data from the source and unlabeled data from the target domain. Our model learns domain agnostic features for tokens using graph convolution networks on token graphs built from labeled and unlabeled data across domains and also preserves local instance level context using a graph attention network on instance graphs. Through our experiments, we showed significant improvements over SOTA methods on $4$ datasets involving binary and multi-label classification. We also implement a BERT based baseline for the first time in this area and show that our method outperforms BERT by a significant margin on average, across datasets. We also showed that our model can retain performance in limited data regime due to effective utilization of unlabeled data. We qualitatively investigated the effects of our model and found GLEN brings features of source and target domains closer. In future, we would investigate the performance of our model from domains other than disaster management.
\ifCLASSOPTIONcaptionsoff
  \newpage
\fi

\clearpage
\newpage


\bibliographystyle{IEEEtran}
\bibliography{IEEEabrv,bare_jrnl_transmag}
\end{document}